\documentclass[sigconf]{acmart}

\usepackage{booktabs}
\usepackage{amsmath}
\usepackage{graphicx}
\usepackage{multirow}
\usepackage{natbib}
\usepackage{comment}
\usepackage{enumitem}
\usepackage{makecell}
\usepackage{tablefootnote}
\usepackage{subcaption}

\setcopyright{rightsretained}

\renewcommand\footnotetextcopyrightpermission[1]{}

\acmDOI{}
\acmISBN{}
\acmConference[PCSC2025]{Philippine Computing Science Congress}{April 2025}{Davao, Philippines}
\acmYear{2025}
\copyrightyear{2025}
\acmArticle{}
\acmPrice{}
\editor{}
\editor{}
\editor{}

\begin{document}
\title{Apparent Age Estimation: Challenges and Outcomes}
\subtitle{A Comparative Analysis on Apparent Age Estimation Methods and Datasets}

\author{Justin Rainier Go}
\email{justin\_rainier\_go@dlsu.edu.ph}
\orcid{0009-0008-2426-4517}
\affiliation{
	\institution{De~La~Salle~University}
	\city{Manila}
	\country{Philippines}
}

\author{Lorenz Bernard Marqueses}
\email{lorenz_marqueses@dlsu.edu.ph}
\orcid{0009-0002-1903-4932}
\affiliation{
	\institution{De~La~Salle~University}
	\city{Manila}
	\country{Philippines}
}

\author{Mikaella Kaye Martinez}
\email{mikaella\_kaye\_martinez@dlsu.edu.ph}
\affiliation{
	\institution{De~La~Salle~University}
	\city{Manila}
	\country{Philippines}
}

\author{John Kevin Patrick Sarmiento}
\email{john\_kevin\_sarmiento@dlsu.edu.ph}
\affiliation{
\institution{De~La~Salle~University}
\city{Manila}
\country{Philippines}
}

\author{Abien Fred Agarap}
\email{abien.agarap@dlsu.edu.ph}
\affiliation{
	\institution{De~La~Salle~University}
	\city{Manila}
	\country{Philippines}
}

\renewcommand{\shortauthors}{Go et al.}

\begin{abstract}
	Apparent age estimation is a valuable tool for business personalization, yet current models frequently exhibit demographic biases.
	We review prior works on the DEX method by applying distribution learning techniques such as Mean-Variance Loss (MVL) and Adaptive Mean-Residue Loss (AMRL), and evaluate them in both accuracy and fairness.
	Using IMDB-WIKI, APPA-REAL, and FairFace, we demonstrate that while AMRL achieves state-of-the-art accuracy, trade-offs between precision and demographic equity persist.
	Despite clear age clustering in UMAP embeddings, our saliency maps indicate inconsistent feature focus across demographics, leading to significant performance degradation for Asian and African American populations.
	We argue that technical improvements alone are insufficient; accurate and fair apparent age estimation requires the integration of localized and diverse datasets, and strict adherence to fairness validation protocols.
\end{abstract}



\maketitle

\section{Introduction and Related Works}

Facial age estimation is a primary focus in computer vision and serves various functions in security, social media, and interactive systems.
A specific variation of this problem that researchers have studied less frequently is apparent age estimation.
This represents the perceived age rather than the actual birth age of a person.
Apparent age estimation is particularly relevant for industries where perceived physical maturity is the central concern such as in the cosmetics industry, and the medical field \cite{swanson_2011_objective,hwang_2010_is}.

More precisely, there are measurable commercial benefits towards personalization including age-related personalization.
Similar studies show that such personalization has yielded a 20--35\% increase in Average Order Value, higher conversion rates due to more relevant recommendations, improved Customer Lifetime Value, and reduced Customer Acquisition Cost through more precise ad targeting~\cite{gupta_2003_customers,hwang_2010_is}.
Given the clear business advantages, apparent age estimation represents a promising research direction.
This problem gained significant prominence during the 2015 ChaLearn Looking at People (CLAP) Challenge \cite{escalera_chalearn_2015}, followed by the introduction of other datasets such as IMDB-WIKI \cite{rothe_2015_dex}, APPA-REAL \cite{agustsson_apparent_2017}, and FairFace \cite{karkkainen_2019_fairface}.
In the following subsections, we briefly discuss related works that were trained and evaluated on the aforementioned datasets.

\subsection{DEX}
\citet{rothe_2015_dex} introduced the Deep Expectation (DEX) method
which uses a VGG-16 architecture pretrained on
the ImageNet dataset as a starting point using a standard cross-entropy loss function.
This was then finetuned on the IMDB-WIKI and CLAP datasets\footnote{ An ensemble of 20 models, each with slightly different finetunes on the CLAP dataset, was used to achieve first place in the 2015 CLAP challenge.
	However,~\citeauthor{rothe_2015_dex} also reported accuracy metrics when using a single CNN.
	We focus on these results for simplicity.
}, although a control group that did not train on IMDB-WIKI was also
tested.
DEX framed its age estimation problem as a classification task using multinomial regression, $E(O) = \sum_{i = 0}^{n=100} y_i o_i$, representing ages from 1 to 101 to produce a predicted age.

\subsection{Mean-Variance Loss}
Unlike DEX, the mean-variance loss \cite{pan_mean-variance_2018} approaches age estimation via distribution learning to capture the correlation between adjacent ages.
Their loss function consists of a mean loss that minimizes the distance between the expected value of the predicted distribution and the ground-truth age, and a variance loss that penalizes the spread of distribution to ensure a sharp and concentrated prediction.
By jointly optimizing these two losses along a softmax cross entropy loss, they demonstrate that their model achieves superior performance across multiple benchmark datasets when compared to DEX.

\subsection{Adaptive Mean-Residue Loss}
The adaptive mean-residue loss \cite{zhao_adaptive_2022} addresses the limitations of conventional distribution-based learning by first estimating a coarse age value and then adaptively calculating a residual value to adjust its prediction towards the actual ground-truth age.
This two-step mechanism allows a network to better handle variances found in facial appearances across different age groups.
With this, the model was able to outperform existing age estimation models at the time with an MAE of 3.61 compared to an MAE of 3.95 through the mean-variance loss.

\subsection{Contributions}

We summarize our main contributions as follows:

\begin{enumerate}
	\item We evaluate model performance trained on IMDB-WIKI using a
	      modified version of the DEX methodology that finetunes on different
	      combinations of the CLAP, APPA-REAL, and FairFace datasets.

	\item We evaluate model performance across demographics
	      (i.e. sex and race), and we identify the degree to
	      which this imbalance can affect the accuracy of both real and apparent age
	      predictions per group.

	\item We explore applications of apparent age estimation in cosmetics, marketing, healthcare, and security to improve personalization and efficiency, and we briefly discuss concerns regarding ethics, privacy, and data governance.

	\item Finally, we lay down some future directions on Philippine facial age estimation preceded with an evaluation on a small-scale dataset.

\end{enumerate}

\section{Exploring Age Estimation datasets}

\subsection{Datasets Description}

\subsubsection{IMDB-WIKIpedia}

\begin{figure}[b]
	\centering
	\includegraphics[width=\linewidth]
	{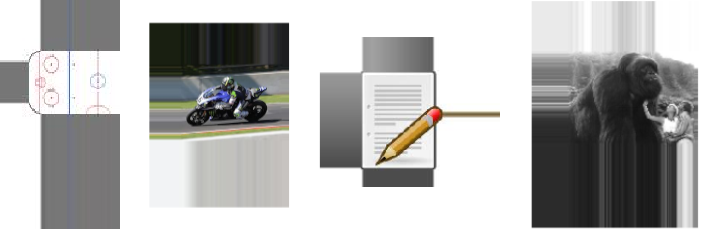}
	\caption{Examples of non-facial images in IMDB-WIKI. \textnormal{Note the presence of pixel-stretching artifacts.}}
	\label{fig:imdb-wiki-dirty}
\end{figure}

This dataset contains images of celebrities scraped from IMDb and Wikipedia and is labeled by offsetting the celebrity's birth date by the timestamp of each image. Initial attempts at model replication using this dataset yielded a higher MAE.
We elected to use a cleaned version of the dataset during training\footnote{ The specific cleaned version of the dataset was retrieved from the following link: \url{github.com/imdeepmind/processed-IMDB-WIKI-dataset}.
} as it appeared that the publicly-available dataset contained anomalous
entries, some of which can be seen in \figurename~\ref{fig:imdb-wiki-dirty}.
The IMDB-WIKI dataset has been noted to have an imbalanced gender ratio of 14:10 in favor of males \cite{puc_analysis_2021}.

\subsubsection{ChaLearn Looking at People (CLAP)}

This was a fairly small dataset created by crowd-sourcing data through an online website \cite{escalera_chalearn_2015}.
It was then augmented by the AgeGuess platform~\cite{jones_ageguess_2019} which collected much of the same data, yielding a final dataset of 4,691 images with a distribution of over 140,000 votes.
Each of these images was annotated with both the real age as well as the mean and variance of their voted apparent ages for use with recognition.
This dataset was the only one in the original DEX pipeline that explicitly annotates its images with apparent age.

\subsubsection{APPA-REAL}

\begin{figure}
	\centering

	\includegraphics[width=\linewidth]
	{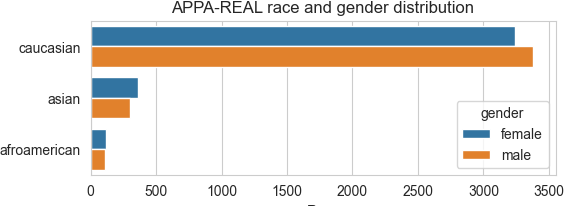}
	\caption{APPA-REAL demographic distribution. \textnormal{The Caucasian group is highly over-represented.}}
	\label{fig:demographics-appa-real}
\end{figure}

\begin{figure}
	\centering
	\includegraphics[width=\linewidth]
	{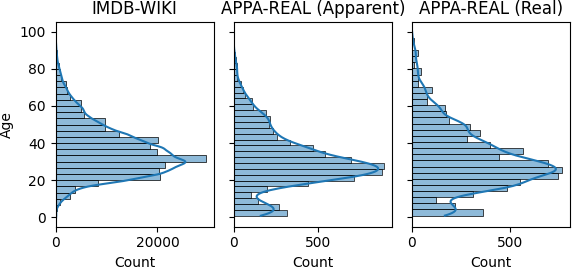}
	\caption{IMDB-WIKI and APPA-REAL age distributions}
	\label{fig:age-dist-IMDB-WIKI}
\end{figure}

The \textit{APPA-REAL} dataset \cite{agustsson_apparent_2017} was the first fairly large dataset to include annotations for both real and apparent age as well as demographic information such as race and gender.
Similar to the CLAP datasets, data was collected through crowdsourcing platforms.
APPA-REAL's real and apparent age distributions (as seen in \figurename~\ref{fig:demographics-appa-real}) are also quite similar to one another with a KL divergence of $D_{KL} \approx 0.01322$.
The APPA-REAL dataset has a similar imbalance to IMDB-WIKI as per \figurename~\ref{fig:demographics-appa-real}, the APPA-REAL dataset is dominated by Caucasian faces, significantly outnumbering Asian and especially African American faces.
\figurename~\ref{fig:age-dist-IMDB-WIKI} shows that despite the size
difference, the real-age distributions across both IMDB-WIKI and APPA-REAL are
somewhat similar, with most of the ages lying within the 20-40 range.

\subsubsection{FairFace}

\begin{figure}
	\centering
	\includegraphics[width=\linewidth]{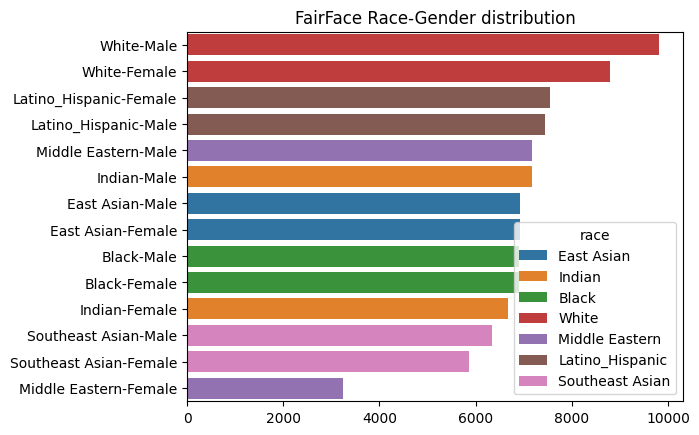}

1	\includegraphics[width=\linewidth]{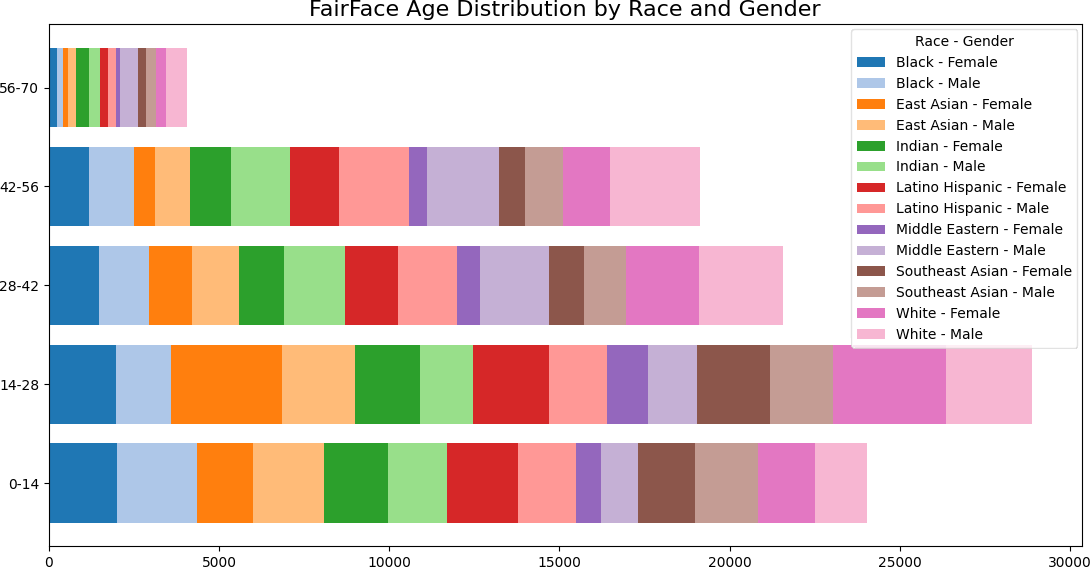}

	\caption{FairFace demographic distribution}
	\label{fig:ff-age}
\end{figure}

The authors claim a fairer distribution of faces in the FairFace \cite{karkkainen_2019_fairface} dataset, though it is still to a lesser extent dominated by Caucasian faces.
It has more race classes, though both male and female groups have roughly as many samples (except for Middle Eastern samples which have much fewer female samples).
However, unlike the other datasets discussed in this paper, FairFace contains no explicit apparent age information; only an age range for each image.
We take the mean of this age range as the ground-truth value during training which may be sufficient for DEX's multinomial regression.
\figurename~\ref{fig:ff-age} demonstrates that each race-gender pair is much
more equally-represented across age groups and that the distribution of age
groups themselves appear to be more balanced.
However, there were still far fewer samples from the older age groups (aged 56 to 70).

\subsection{Evaluation Metrics}\label{sec:model-evaluation-criteria}

We evaluate models through their mean absolute error (MAE) values.
Additionally, we use the $\epsilon$-error to account for the degree of uncertainty when estimating the apparent age.
Formally defined as $1 - \exp\left({-\frac{(x-\mu)^2}{2 \sigma^2}}\right)$, the $\epsilon$-error fits a normal distribution based on the mean and standard deviation of the collected user guesses for a given image~\cite{escalera_chalearn_2015}.
Results shown in Table~\ref{tab:performance} are presented as they were reported in their respective papers, while results from our reproduction can be found in \tablename s~\ref{tab:mae} and \ref{tab:mae_race} -- all in Section \ref{sect:evaluation}.

\newpage
\subsection{Modelling}

We finetuned on the following combinations of datasets:

\begin{enumerate}
	\item IMDB-WIKI;
	\item IMDB-WIKI, then CLAP;
	\item IMDB-WIKI, then APPA-REAL;
	\item IMDB-WIKI, then FairFace;
	\item IMDB-WIKI, then FairFace, then CLAP; and
	\item IMDB-WIKI, then FairFace, then APPA-REAL.
\end{enumerate}

At the same time, we also attempt replacing the cross-entropy loss (CEL) function of the original DEX method with more recent loss functions, specifically, mean-variance loss and adaptive mean-residue loss.
With six combinations of datasets and three choices of loss functions, we end up with 18 different models to evaluate.
To evaluate the models, we continue to use the model evaluation metrics in Section~\ref{sec:model-evaluation-criteria}, and generate UMAP embeddings and cosine similarity graphs.
Additionally, we produce saliency maps to explain which regions of each image held the most weight when predicting the apparent ages of people, grouped by race and gender, with one image for each age group.
The representative image was the closest to the midpoint of each age group.

\subsection{Bias and fairness assessment}

Imbalances in sample distribution across race and gender can be seen most strongly in the APPA-REAL dataset.
As we want to measure the variance in accuracy of the overall models based on training data demographics, we kept the datasets as-is, although we hypothesized that the models finetuned on FairFace would experience a smaller degree of racial and sexual sampling biases due to its greater diversity.

\subsection{Software/Hardware configuration}

\citet{rothe_2015_dex}'s software setup utilized Caffe, an open-source deep
learning framework.
Our replication of the model and the further variants use PyTorch Lightning with Python versions $\geq$~\texttt{3.11}, with our repositories using \texttt{uv} for project management.
Due to computational and time constraints, we used Kaggle's NVIDIA P100 for training and finetuning on IMDB-WIKI and FairFace while we used NVIDIA RTX 3060 and 4060 GPUs on CLAP and APPA-REAL respectively.

\section{Evaluating Age Estimation models}\label{sect:evaluation}
We present our validation of the reported performance trend of previous works, namely DEX \cite{rothe_2015_dex}, MVL \cite{pan_mean-variance_2018}, and AMRL \cite{zhao_adaptive_2022}\footnote{ The code used to produce our results are stored in a group of repositories hosted in GitLab, accessible through the following link:

	\url{https://gitlab.com/data100-s12-group7/imdb-wiki/}.
}.
Our results corroborate the general trend of the newer approaches outperforming earlier ones.

\begin{table}
	\centering
	\small
	\caption{Comparison of DEX, MVL, and AMRL loss as reported by each paper}
	\label{tab:performance}
	\begin{tabular}{l|cc|cc|cc}
		\hline
		                                                   & \multicolumn{2}{c|}{\cite{rothe_2015_dex}} &
		\multicolumn{2}{c|}{\cite{pan_mean-variance_2018}} &
		\multicolumn{2}{c}{\cite{zhao_adaptive_2022}}                                                                                                                                                \\
		                                                   & \multicolumn{2}{c|}{CLAP 2015}             & \multicolumn{2}{c|}{FG-NET, CLAP
		2016}                                              &
		\multicolumn{2}{c}{FG-NET, CLAP
			2016}
		\\
		                                                   & MAE                                        & $\epsilon$                       & MAE           & $\epsilon$ & MAE           & $\epsilon$ \\
		\hline

		DEX                                                & \textbf{3.22}                              & 0.28                             & 3.09          & -          & 4.63          & -          \\
		MVL                                                & -                                          & -                                & \textbf{2.68} & 0.29       & 3.95          & 0.40       \\
		AMRL                                               & -                                          & -                                & -             & -          & \textbf{3.61} & 0.39       \\
		\hline
	\end{tabular}
\end{table}

\begin{table}
	\centering
	\small
	\caption{MAE and $\epsilon$-error results. \textnormal{These results do not
			yet include the FairFace finetuning step as we attempt to reproduce the
			results of \cite{rothe_2015_dex} while comparing CLAP and APPA-REAL.
			Star symbol ($^*$) indicates Caffe model from \cite{rothe_2015_dex}.
			Plus symbol ($^+$) indicates our own implementation.
		}}
	\label{tab:mae}

	\begin{tabular}{l|l|ll|l}
		\hline
		                              &                             & \multicolumn{2}{c|}{MAE} &                 \\
		Testing set                   & Model                       &
		Appa.*                        & Real                        & $\epsilon$-error                           \\
		\hline
		\multirow{6}{*}{APPA-REAL}    & CLAP$^{+}$                  &
		12.57                         & 11.20                       & 0.88                                       \\
		                              & IMDB-WIKI + CLAP$^{*}$      & 5.18                     & 6.84          &
		0.38                                                                                                     \\
		                              & IMDB-WIKI$^{*}$             & 6.96                     & 7.89          &
		0.49                                                                                                     \\
		                              & IMDB-WIKI + APPA-REAL$^{+}$ & \textbf{4.32}            & \textbf{6.26} &
		\textbf{0.34}                                                                                            \\
		                              & IMDB-WIKI + CLAP$^{+}$      & 5.51                     & 7.04          &
		0.41                                                                                                     \\
		                              & IMDB-WIKI$^{+}$             & 5.93                     & 7.19          &
		0.43                                                                                                     \\
		\hline
		\multirow{6}{*}{CLAP} & CLAP$^{+}$                  &
		-                             & 6.64                        & 0.74                                       \\
		                              & IMDB-WIKI + CLAP$^{*}$      & -                        & \textit{3.25} &
		\textit{0.28}                                                                                            \\
		                              & IMDB-WIKI$^{*}$             & -                        & 5.66          &
		0.49                                                                                                     \\
		                              & IMDB-WIKI + APPA-REAL$^{+}$ & -                        & 3.77          &
		0.34                                                                                                     \\
		                              & IMDB-WIKI + CLAP$^{+}$      & -                        & \textbf{3.57} &
		\textbf{0.32}                                                                                            \\
		                              & IMDB-WIKI$^{+}$             & -                        & 4.90          &
		0.44                                                                                                     \\
		\hline
	\end{tabular}
\end{table}

\begin{table}
	\centering
	\small
	\caption{MAE and $\epsilon$-error results
		per model on APPA-REAL, grouped by race and gender, ordered by MAE
		(apparent age). \textnormal{We present results from the 5
			best-performing models on apparent age.
			The MAE values for the best and worst groups per model are shown in bold.
		}}
	\label{tab:mae_race}

	\begin{tabular}{l|ll|rr|r}
		\toprule
		                                 &                               &               & \multicolumn{2}{c|}{MAE} &                   \\
		Model                            & Race                          & Gender        & Apparent                 & Real & $\epsilon$ \\
		\midrule
		\multirowcell{6}{AMRL:                                                                                                          \\ IMDB-WIKI\\ + APPA-REAL}                     &
		\multirow[t]{2}{*}{African American} & Female                        & 2.93          & 5.15                     & 0.27              \\
		                                 &                               & Male          & \textbf{2.92}            & 6.45 & 0.23       \\
		\cline{2-6}
		                                 & \multirow[t]{2}{*}{Asian}     & Female        & \textbf{4.33}            & 6.41 & 0.38       \\
		                                 &                               & Male          & 3.77                     & 4.92 & 0.36       \\
		\cline{2-6}
		                                 & \multirow[t]{2}{*}{Caucasian} & Female        & 4.08                     & 6.14 & 0.32       \\
		                                 &                               & Male          & 3.53                     & 5.00 & 0.28       \\
		\cline{1-6} \cline{2-6}
		\multirowcell{6}{AMRL:                                                                                                          \\ IMDB-WIKI\\ + FairFace\\ + APPAREAL}                      &
		\multirow[t]{2}{*}{African American} & Female                        & 4.60          & 6.24                     & 0.36              \\
		                                 &                               & Male          & 3.85                     & 7.14 & 0.31       \\
		\cline{2-6}
		                                 & \multirow[t]{2}{*}{Asian}     & Female        & \textbf{4.62}            & 6.91 & 0.40       \\
		                                 &                               & Male          & \textbf{3.40}            & 5.34 & 0.33       \\
		\cline{2-6}
		                                 & \multirow[t]{2}{*}{Caucasian} & Female        & 4.26                     & 6.45 & 0.34       \\
		                                 &                               & Male          & 3.82                     & 5.17 & 0.31       \\
		\cline{1-6} \cline{2-6}
		\multirowcell{6}{CEL:                                                                                                           \\ IMDB-WIKI\\ + FairFace\\ + APPA-REAL}                     &
		\multirow[t]{2}{*}{African American} & Female                        & \textbf{4.62} & 6.21                     & 0.37              \\
		                                 &                               & Male          & \textbf{2.92}            & 6.38 & 0.22       \\
		\cline{2-6}
		                                 & \multirow[t]{2}{*}{Asian}     & Female        & 4.53                     & 6.84 & 0.36       \\
		                                 &                               & Male          & 4.46                     & 6.17 & 0.40       \\
		\cline{2-6}
		                                 & \multirow[t]{2}{*}{Caucasian} & Female        & 4.48                     & 6.85 & 0.35       \\
		                                 &                               & Male          & 4.18                     & 5.50 & 0.33       \\
		\cline{1-6} \cline{2-6}
		\multirowcell{6}{MVL:                                                                                                           \\ IMDB-WIKI\\ + APPA-REAL}                     &
		\multirow[t]{2}{*}{African American} & Female                        & 3.30          & 4.17                     & 0.27              \\
		                                 &                               & Male          & \textbf{2.89}            & 5.85 & 0.24       \\
		\cline{2-6}
		                                 & \multirow[t]{2}{*}{Asian}     & Female        & \textbf{4.65}            & 7.24 & 0.36       \\
		                                 &                               & Male          & 4.17                     & 5.73 & 0.42       \\
		\cline{2-6}
		                                 & \multirow[t]{2}{*}{Caucasian} & Female        & 4.34                     & 6.11 & 0.34       \\
		                                 &                               & Male          & 3.49                     & 4.91 & 0.29       \\
		\cline{1-6} \cline{2-6}
		\multirowcell{6}{MVL:                                                                                                           \\ IMDB-WIKI\\ + FairFace\\ + APPA-REAL}                     &
		\multirow[t]{2}{*}{African American} & Female                        & \textbf{3.22} & 4.81                     & 0.26              \\
		                                 &                               & Male          & 3.58                     & 6.77 & 0.27       \\
		\cline{2-6}
		                                 & \multirow[t]{2}{*}{Asian}     & Female        & \textbf{4.65}            & 6.77 & 0.41       \\
		                                 &                               & Male          & 3.89                     & 5.50 & 0.37       \\
		\cline{2-6}
		                                 & \multirow[t]{2}{*}{Caucasian} & Female        & 4.06                     & 6.26 & 0.31       \\
		                                 &                               & Male          & 3.53                     & 5.03 & 0.29       \\
		\bottomrule
	\end{tabular}
\end{table}

Table \ref{tab:performance} shows lifted results from literature \cite{rothe_2015_dex, pan_mean-variance_2018, zhao_adaptive_2022}.
Table \ref{tab:mae} summarizes the performances of the recreated models on both the APPA-REAL and CLAP testing datasets.
Table \ref{tab:mae_race} shows how this performance varies across different groups of race and gender.
The evaluation reveals performance disparities across demographic intersections.
The models exhibit the highest error rates on African American and Asian female subjects while they perform better on African American male counterparts.
Interestingly, this latter group presents the most substantial divergence between apparent and real age.
These results suggest that the performance gap is a direct consequence of the dataset distribution which is heavily skewed towards male subjects and thus lacks the necessary representation to ensure fairer performance across demographics.

Figures \ref{fig:cel_embeddings} and \ref{fig:mvl_amrl_embeddings} show UMAP-reduced image embeddings for three models of interest when tested on the APPA-REAL dataset: the model trained only on the CLAP train set, the model trained on only IMDB-WIKI, and the model trained on IMDB-WIKI and finetuned on the APPA-REAL train set.
The model trained solely on the CLAP dataset performed the worst, as it learned considerably fewer facial representations compared to other models.
However, there still appear to be some prominent clusters in the reduced embedding space, particularly with younger ages (purple hues).

The model finetuned on the APPA-REAL training set had better results, showing clearer distinction for both younger (yellow-green hues) and older ages.
On the other hand, it performs worse with ``middle'' age groups, despite the age distribution; attributable to strong feature overlap between individuals in these age groups.
With APPA-REAL, we find the similar trend but with visibly better clustering.
We acknowledge that these are qualitative results, and we intend to present quantitative metrics (e.g. mean discrepancies between age groups) to support these findings in a future work.

\begin{figure}
	\centering

	\includegraphics[width=\linewidth]{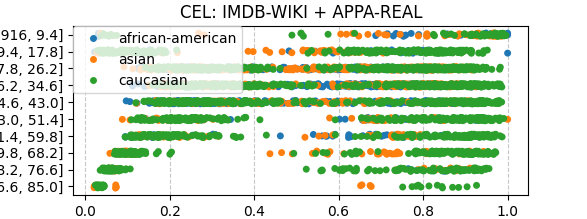}

	\caption{Cosine similarity of each image with average embeddings for age, grouped by race}
	\label{fig:combined_cosine_appareal}
\end{figure}

Figure \ref{fig:combined_cosine_appareal} shows the cosine similarity of sample images with the average embeddings for its respective age group for the model finetuned on only APPA-REAL.
Notably, despite good model performance, the distribution of similarity values appears to be spread out.
We posit this is because while the classifier can select the best age class for an image, this does not guarantee that it is highly similar to the average embeddings for that class.

\subsection{Model Performance}

\begin{table*}[h]
	\small
	\centering
	\caption{MAE results per model ordered by MAE (apparent). \textnormal{We
			show the top 10 best-performing models in terms of apparent age MAE.
			Best model results are shown in bold.
		}}
	\label{tab:mae_}
	\begin{tabular}{l|rrrrrr}
		\toprule
		                                                            & \multicolumn{2}{r}{MAE (apparent)} & \multicolumn{2}{r}{MAE (real)} &
		\multicolumn{2}{r}{$\epsilon$}                                                                                                                               \\
		Model                                                       & mean                               & std                            & mean & std  & mean & std \\
		\midrule
		Adaptive mean-residue loss: IMDB-WIKI + APPA-REAL           & \textbf{3.59}                      &
		0.58                                                        & 5.68
		                                                            & 0.73                               & 0.31                           & 0.06                     \\
		Mean-variance loss: IMDB-WIKI + APPA-REAL                   & 3.81                               & 0.68                           & 5.67 & 1.05 &
		0.32                                                        & 0.07                                                                                           \\
		Mean-variance loss: IMDB-WIKI + FairFace + APPA-REAL        & 3.82                               & 0.50                           &
		5.86                                                        & 0.86                               & 0.32                           & 0.06                     \\
		Adaptive mean-residue loss: IMDB-WIKI + FairFace + APPAREAL & 4.09                               &
		\textbf{0.49}                                               & 6.21                               & 0.81                           & 0.34 & 0.04              \\
		Cross-entropy loss: IMDB-WIKI + FairFace + APPA-REAL        & 4.20                               & 0.64                           &
		6.32                                                        & 0.50                               & 0.34                           & 0.06                     \\
		Cross-entropy loss: IMDB-WIKI + APPA-REAL                   & 4.46                               & 0.54                           & 6.47 & 0.54 &
		0.37                                                        & 0.06                                                                                           \\
		Adaptive mean-residue loss: IMDB-WIKI + CHALEARN            & 4.98                               & 0.67                           & 6.70 &
		0.76                                                        & 0.40                               & 0.05                                                      \\
		Original Caffe model on IMDB-WIKI + CHALEARN                & 4.99                               & 0.57                           & 6.90 &
		0.95                                                        & 0.39                               & 0.02                                                      \\
		Mean-variance loss: IMDB-WIKI + CHALEARN                    & 5.03                               & 0.89                           & 6.78 & 0.84 &
		0.39                                                        & 0.06                                                                                           \\
		Mean-variance loss: IMDB-WIKI + FairFace + CHALEARN         & 5.07                               & 0.67                           &
		7.11                                                        & 0.72                               & 0.41                           & 0.07                     \\
		Mean-variance loss: IMDB-WIKI + FairFace                    & 5.14                               & 0.61                           & 6.70 & 0.75 &
		0.42                                                        & 0.06                                                                                           \\

		\bottomrule
	\end{tabular}
	\parbox{\linewidth}{\raggedright \footnotesize *``std'' column shows the standard deviation of the MAE values
		when grouped by race and gender; this summarizes variance in model performance among race-gender pairs. }
\end{table*}

Table \ref{tab:mae_} illustrated that the model utilizing the adaptive mean-residue loss achieves the highest precision when evaluated against the APPA-REAL test set.
Specifically, the configuration involving pretraining on IMDB-Wiki followed by finetuning on APPA-REAL yielded the lowest MAE.
In contrast, the model incorporating an immediate finetuning stage on the FairFace dataset before the APPA-REAL training exhibited the lowest variance in performance across demographics.
These findings suggest that while the inclusion of FairFace dataset does not necessarily improve overall performance, it serves as a paramount step for introducing demographic equity by ensuring more consistent results across diverse populations.

\subsection{Visualizing Facial Age Representations}

\begin{figure}
	\centering

	\includegraphics[width=\linewidth]{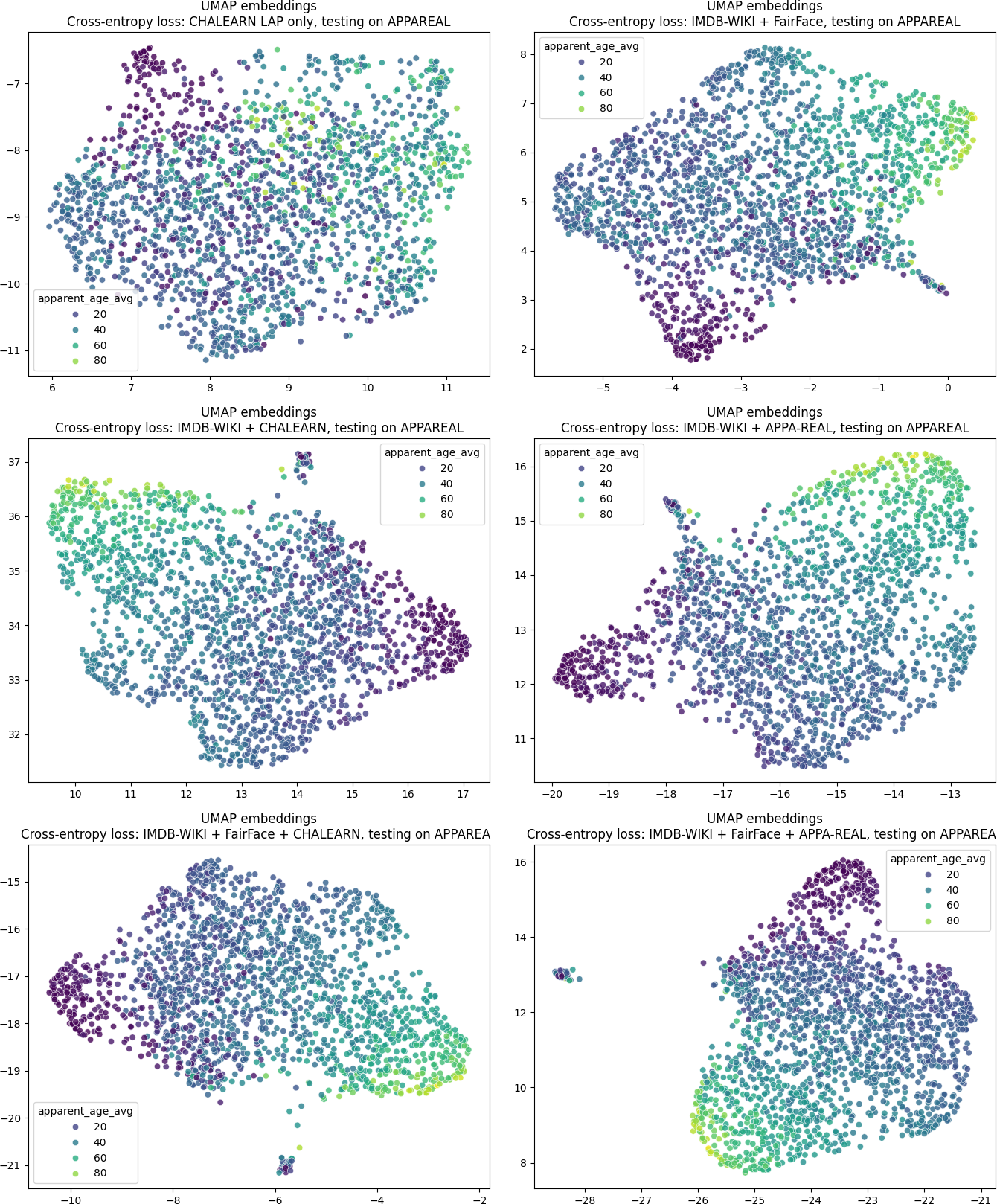}

	\caption[UMAP-reduced embeddings]{
		UMAP-reduced embeddings on APPA-REAL test set using different finetuning combinations with cross-entropy loss.
        \textnormal{Left column, top to bottom: CLAP only, IMDB-WIKI + CLAP, and IMDB-WIKI + FairFace + CLAP. Right column replaces CLAP with APPA-REAL.}
	}
	\label{fig:cel_embeddings}
\end{figure}

\begin{figure}
	\centering

	\includegraphics[width=\linewidth]{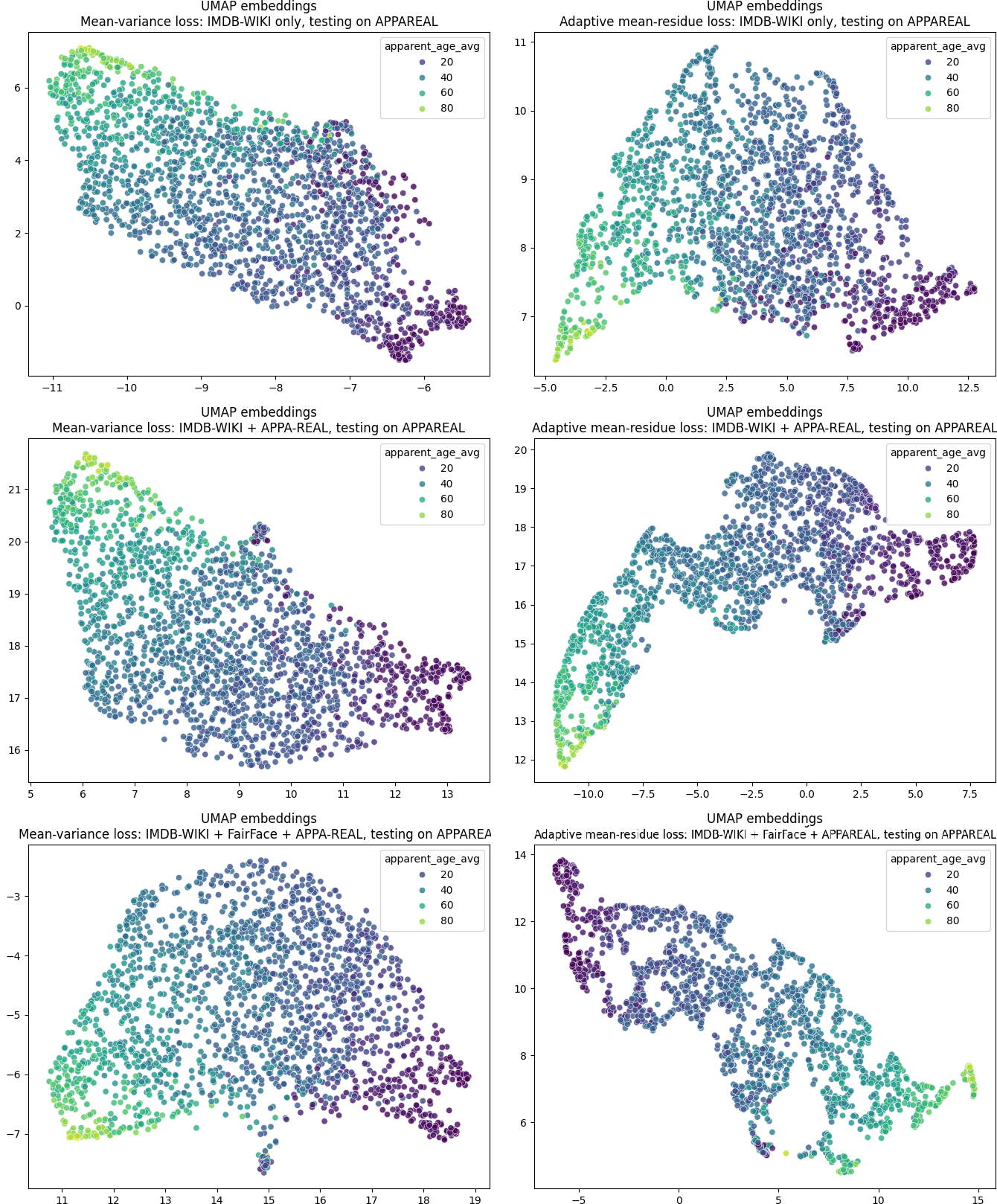}

	\caption[UMAP-reduced embeddings]{
		UMAP-reduced embeddings using different finetuning combinations with
		MVL (left column) and AMRL (right column).
		\textnormal{
            Follows the same progression of APPA-REAL-based models as in \autoref{fig:cel_embeddings}.
			AMRL with FairFace finetuning (bottom right) appears to result in
			stronger clustering of age ranges.
		}
	}
	\label{fig:mvl_amrl_embeddings}
\end{figure}

Our experimental results indicate that while all models successfully identify certain demographic clusters, the models trained with the adaptive mean-residue loss yield the best clustering of embeddings.
As illustrated in Figure \ref{fig:mvl_amrl_embeddings}, the embedding space reveals two distinct and well-defined clusters.
This visualization mirrors patterns observed in Figure \ref{fig:cel_embeddings} where the model demonstrates a high proficiency in distinguishing ages at the extreme ends of the spectrum particularly for younger subjects.
Specifically, the AMRL embeddings in Figure \ref{fig:mvl_amrl_embeddings} highlight a clearly distinguishable purple ``island'' within the latent space which represents a highly concentrated and separated grouping of specific age features.

\subsection{Recognizing Facial Features}

\begin{figure}[b]

	\begin{subfigure}[b]{\linewidth}
		\centering

		\includegraphics[width=\linewidth]{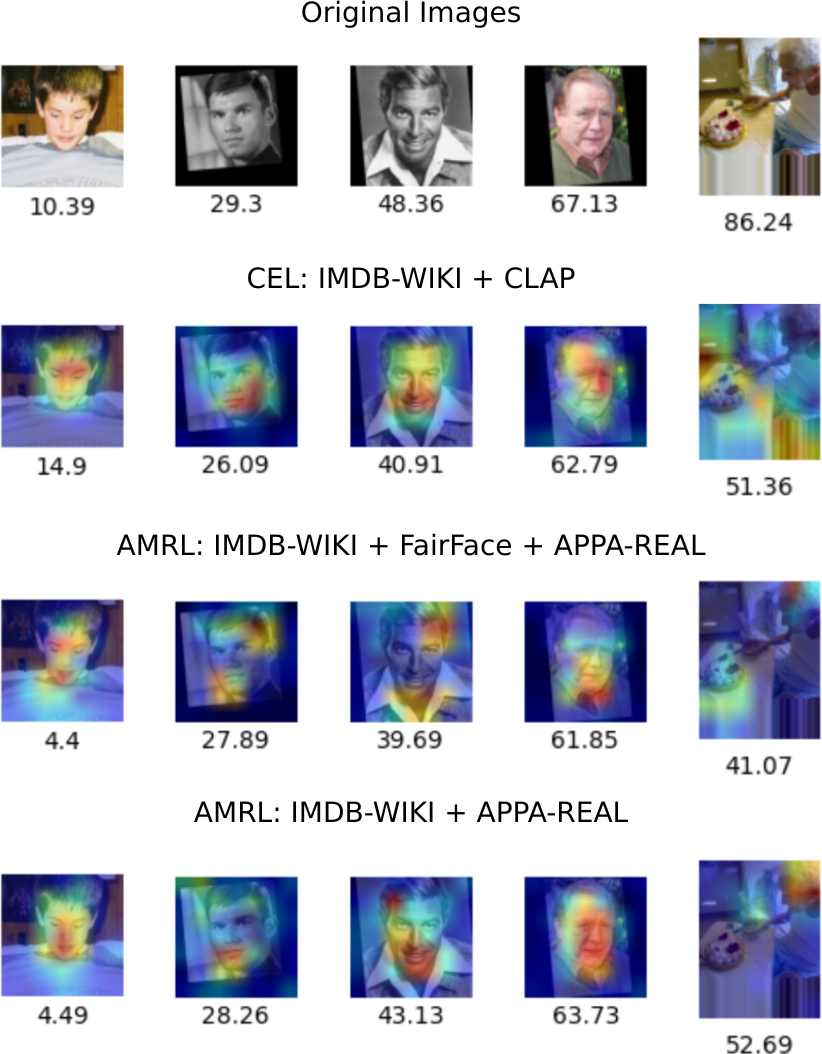}
		\caption{Caucasian Males}
		\label{fig:saliency-cm}
	\end{subfigure}

	\begin{subfigure}[b]{\linewidth}
		\centering
		\includegraphics[width=\linewidth]{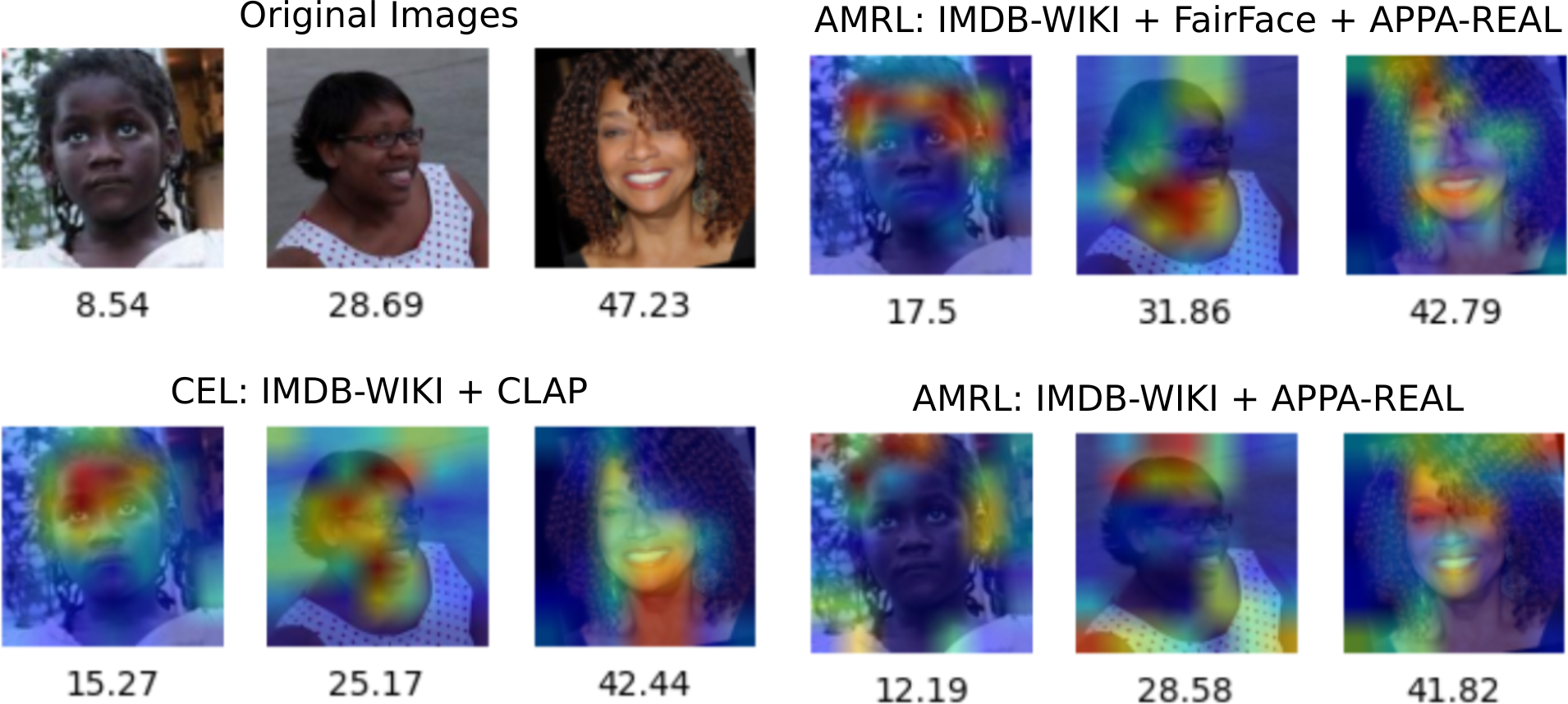}
		\caption{African American Females}
		\label{fig:saliency-af}
	\end{subfigure}

	\caption{Saliency maps indicate feature focus inconsistencies}
	\label{fig:saliency}

\end{figure}

Figure \ref{fig:saliency} presents saliency maps overlaid on input images along the ground-truth and estimated apparent ages to identify the regions that contribute the most to the final model prediction.
In these visualizations, redder areas denote higher importance for the decisions.
Across the various evaluated models, the features influencing the age estimation remain largely consistent with the most critical regions concentrated on the center of their faces.
However, when analyzing specific demographic groups, the model focuses more on peripheral areas such as the forehead or neck.
Furthermore, Figure \ref{fig:saliency-cm} demonstrates that the model can still produce incorrect estimations for younger faces even up to a decade despite highlighting the appropriate facial features.
We acknowledge that this is fairly qualitative analysis, and leaves room for a more quantitative approach in a future work.

\subsection{Facial Age Group Similarities}

\begin{figure}
	\centering

	\includegraphics[width=\linewidth]{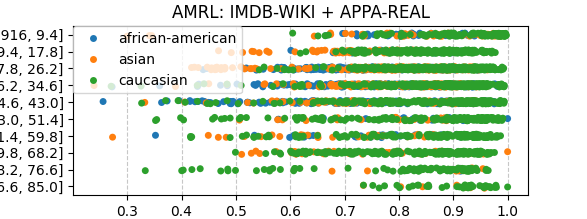}

	\caption{Cosine similarity of each image with average embeddings for age (trained via AMRL, on IMDB-WIKI, then APPA-REAL, grouped by race)}
	\label{fig:combined_cosine_appareal_amrl}
\end{figure}

When finetuning via AMRL (as in \figurename~\ref{fig:combined_cosine_appareal_amrl}), cosine similarity is much more concentrated towards 1.0 compared to other methods (e.g. as seen in \figurename~\ref{fig:combined_cosine_appareal}, whose cosine similarities are more spread out despite the accuracy).
This is possibly because the adaptive mean-residue loss function doesn't neccessarily necessitate that predicted ages are highly similar to the target age, only that they are most similar to it.
Nevertheless, it appears that AMRL-finetuned models appear to achieve both goals.

\subsection{Localized Age Estimation}
\begin{figure}
	\centering

	\includegraphics[width=\linewidth]{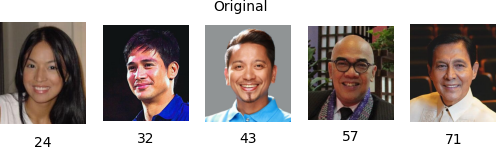}
	\includegraphics[width=\linewidth]{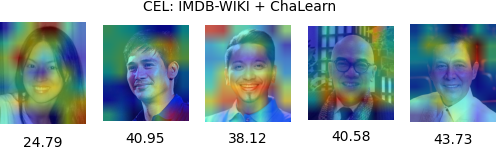}
	\includegraphics[width=\linewidth]{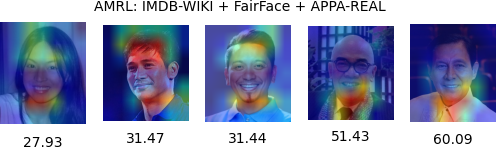}
	\includegraphics[width=\linewidth]{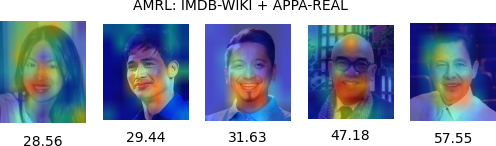}

	\caption{Saliency maps on Filipino celebrities similarly show inconsistent feature focus}
	\label{fig:saliency-filipino}
\end{figure}

We also evaluated our two leading AMRL models and our CEL replication against a self-annotated dataset of forty Filipino celebrity images.
As shown in Table \ref{tab:mae_filipino}, the model finetuned on the FairFace dataset performed the best, mirroring our main findings.
Despite the superior performance of the AMRL model over CEL and MVL models, Figure \ref{fig:saliency-filipino} reveals similar anomalies where the model focuses on isolated or non-essential facial regions.

\begin{table}
	\centering
	\caption{MAE results on Filipino celebrity dataset}
	\label{tab:mae_filipino}
	\begin{tabular}{l|r}
		\hline
		Model                                  & MAE   \\
		\hline
		CEL: IMDB-WIKI + ChaLearn              & 10.05 \\
		AMRL: IMDB-WIKI + FairFace + APPA-REAL & 6.82  \\
		AMRL: IMDB-WIKI + APPA-REAL            & 7.05  \\
		\hline
	\end{tabular}
\end{table}

\newpage
\section{Business Implications}
Apparent age estimation offers substantial commercial utility in sectors requiring advanced personalization and consumer profiling.
In the cosmetics industry, brands utilize facial analytics to assess a consumer's perceived appearance, facilitating data driven skincare recommendations tailored to specific aging markers.
Furthermore, in clinical and dermatological contexts, perceived age analysis supports the identification of lifestyle-related aging, dermatological pathologies, and stress-induced changes.
A notable example is the FaceAge system which utilizes facial data to estimate biological age and assist in health prognostication \cite{Bontempi2025-xm}.

In Know-Your-Customer (KYC) systems, apparent age estimation enhances identity verification by validating whether a subject's facial features align with the age bracket indicated on their credentials.
This provides an additional layer of fraud detection for financial institutions, fintech firms, and digital platforms.
Furthermore, integrating this technology helps prevent minors from accessing age-restricted services, thereby strengthening institutional compliance and security.

\subsection{Inclusivity and demographic representativeness}
Several models exhibit substantial performance disparities resulting from dataset imbalances.
Specifically, Asian and African American female cohorts consistently yield the highest MAE, whereas male demographics achieve the lowest error rates.
Within business applications such as KYC, these inconsistencies heighten the risk of false fraud flags, operational delays, and inequitable account rejections.
The overrepresentation of Caucasian subjects in training data skews the learned representations, leading to systematic underperformance for other racial and gender groups.

\subsection{Ethical, privacy, data governance issues}
Deploying age estimation models within the Philippine context introduces critical challenges regarding bias, privacy, and transparency.
Primarily, models trained predominantly on Western datasets often fail to accurately interpret Southeast Asian facial features.
These technical inaccuracies risk reinforcing historical prejudices concerning skin tone and beauty standards, particularly within the cosmetics industry which can lead to a significant loss of brand trust.

Furthermore, the Philippine Data Privacy Act of 2012 classifies facial images as sensitive personal information.
Many organizations utilize cloud infrastructures that may lack the robust security protocols necessary to prevent data breaches.
The frequent lack of transparency regarding data storage and third party access remains a significant legal and ethical hurdle.

Finally, inadequate data governance presents substantial legal risks.
Failure to conduct mandatory Privacy Impact Assessments constitutes a violation of existing regulations.
Moreover, the utilization of improperly scraped data compromises model integrity and heightens the potential for harm to Filipino users.

\subsection{Strategic recommendations for organizations}
To mitigate these concerns and avoid legal repercussions, we believe organizations must prioritize the development and integration of localized datasets.
It is essential to develop and validate models using facial data representative of Filipino and Southeast Asian populations.
This ensures that algorithmic performance remains equitable regardless of skin tone or age.

Such models should be integrated directly into existing infrastructures, including consumer applications and financial security systems.
Within the skincare industry, this localization facilitates more precise and personalized product recommendations.
For the banking sector, it enhances fraud detection capabilities while ensuring compliance with Bangko Sentral ng Pilipinas regulations.
This strategic approach enables companies to monitor model performance effectively, ensuring sustained ethical standards and technical accuracy.

\section{Conclusion}
The model trained on IMDB-WIKI followed by APPA-REAL utilizing the AMRL loss performed the best in apparent age estimation.
Conversely, the configuration that incorporated FairFace in its finetuning demonstrated the lowest variance across racial and gender demographics.
Overall, the AMRL method proved most effective for this task.
UMAP visualizations of AMRL embeddings showed well-defined clusters, while saliency gradients indicated a more consistent focus on facial regions compared to alternative methods.
Additionally, cosine similarity analysis for corresponding apparent and real age embeddings tended toward unity under this approach.
While this technology offers diverse commercial potential, its adoption remains constrained by pervasive demographic imbalances favoring Caucasian populations, alongside critical ethical, privacy, and data governance challenges.

\section{Future works}

In addition to the quantitative metrics we intend to pursue in a future study,
we lay down some of the more specific and broader research we shall pursue:

\begin{enumerate}
	\item \textbf{Contrastive Learning for Few-shot Representation}.
	      We propose investigating contrastive learning techniques to engender age estimation models with few-shot capabilities, particularly for underrepresented demographics.
	      Current models often rely on representations learned from datasets where East Asian populations like Taiwan, Korean, or Japan are better represented than Austronesian groups.
	      Research into how these learned features generalize to Filipino faces could reveal critical performance gaps.
	      Contrastive objectives may provide a more robust way to learn distinct facial features from limited samples, thus ensuring that model performance remains equitable across ethnic groups.

	\item \textbf{Longitudinal Filipino Celebrity Dataset}.
	      A significant hurdle in age estimation research is the scarcity of localized longitudinal data.
	      We intend to extend our limited Filipino celebrity age estimation dataset to a cross-age dataset \'a la \citet{chen2014cross}.
	      By creating such a dataset, we can better observe unique physiological aging patterns within the Filipino population.
	      This resource would then help facilitate the development of models that are better calibrated to local features rather than relying on global averages that may not accurately reflect regional facial representations.

	\item \textbf{Optimization for low-resourced compute}.
	      Finally, we suggest exploring contrastive learning to improve the specialization of individual components within a mixture-of-experts \cite{jacobs1991adaptive} architecture.
	      In this framework, different experts can be trained to specialize in specific age brackets or demographic features.
	      Beyond improving accuracy through this specialization, this approach should be evaluated for its potential to enhance inference speed and compute resource utilization.
	      By selectively activating only the most relevant experts for a given input, we can maintain a high throughput with a highly accurate age estimation model.

\end{enumerate}

\bibliographystyle{ACM-Reference-Format}
\bibliography{bibfile}

@inproceedings{rothe_2015_dex,
  author    = {Rothe, Rasmus and Timofte, Radu and Van Gool, Luc},
  booktitle = {2015 IEEE International Conference on Computer Vision Workshop (ICCVW)},
  title     = {DEX: Deep EXpectation of Apparent Age from a Single Image},
  year      = {2015},
  volume    = {},
  number    = {},
  pages     = {252-257},
  doi       = {10.1109/ICCVW.2015.41}
}

@inproceedings{pan_mean-variance_2018,
  address   = {Salt Lake City, UT, USA},
  title     = {Mean-{Variance} {Loss} for {Deep} {Age} {Estimation} from a {Face}},
  isbn      = {978-1-5386-6420-9},
  url       = {https://ieeexplore.ieee.org/document/8578652/},
  doi       = {10.1109/CVPR.2018.00554},
  urldate   = {2025-10-23},
  booktitle = {2018 {IEEE}/{CVF} {Conference} on {Computer} {Vision} and {Pattern} {Recognition}},
  publisher = {IEEE},
  author    = {Pan, Hongyu and Han, Hu and Shan, Shiguang and Chen, Xilin},
  month     = jun,
  year      = {2018},
  pages     = {5285--5294}
}

@inproceedings{zhao_adaptive_2022,
  address   = {Taipei, Taiwan},
  title     = {Adaptive {Mean}-{Residue} {Loss} for {Robust} {Facial} {Age} {Estimation}},
  copyright = {https://doi.org/10.15223/policy-029},
  isbn      = {978-1-6654-8563-0},
  url       = {https://ieeexplore.ieee.org/document/9859703/},
  doi       = {10.1109/ICME52920.2022.9859703},
  urldate   = {2025-10-23},
  booktitle = {2022 {IEEE} {International} {Conference} on {Multimedia} and {Expo} ({ICME})},
  publisher = {IEEE},
  author    = {Zhao, Ziyuan and Qian, Peisheng and Hou, Yubo and Zeng, Zeng},
  month     = jul,
  year      = {2022},
  pages     = {1--6}
}

@inproceedings{escalera_chalearn_2015,
  address    = {Santiago, Chile},
  title      = {{ChaLearn} {Looking} at {People} 2015: {Apparent} {Age} and
                {Cultural} {Event} {Recognition} datasets and results},
  isbn       = {978-1-4673-9711-7},
  shorttitle = {{ChaLearn} {Looking} at {People} 2015},
  url        = {http://ieeexplore.ieee.org/document/7406389/},
  doi        = {10.1109/ICCVW.2015.40},
  urldate    = {2025-10-24},
  booktitle  = {2015 {IEEE} {International} {Conference} on {Computer}
                {Vision} {Workshop} ({ICCVW})},
  publisher  = {IEEE},
  author     = {Escalera, Sergio and Fabian, Junior and Pardo, Pablo and Baro,
                Xavier and Gonzalez, Jordi and Escalante, Hugo J. and Misevic, Dusan and
                Steiner, Ulrich and Guyon, Isabelle},
  month      = dec,
  year       = {2015},
  pages      = {243--251}
}

@article{jones_ageguess_2019,
  title    = {The {AgeGuess} database, an open online resource on chronological
              and perceived ages of people aged 5–100},
  volume   = {6},
  issn     = {2052-4463},
  url      = {https://www.nature.com/articles/s41597-019-0245-9},
  doi      = {10.1038/s41597-019-0245-9},
  language = {en},
  number   = {1},
  urldate  = {2025-10-24},
  journal  = {Scientific Data},
  author   = {Jones, Julia A. Barthold and Nash, Ulrik W. and Vieillefont,
              Julien and Christensen, Kaare and Misevic, Dusan and Steiner, Ulrich K.},
  month    = oct,
  year     = {2019},
  pages    = {246}
}

@misc{karkkainen_2019_fairface,
  title         = {FairFace: Face Attribute Dataset for Balanced Race, Gender, and Age},
  author        = {Kimmo Kärkkäinen and Jungseock Joo},
  year          = {2019},
  eprint        = {1908.04913},
  archiveprefix = {arXiv},
  primaryclass  = {cs.CV},
  url           = {https://arxiv.org/abs/1908.04913}
}

@inproceedings{agustsson_apparent_2017,
  address   = {Washington, DC, DC, USA},
  title     = {Apparent and {Real} {Age} {Estimation} in {Still} {Images} with
               {Deep} {Residual} {Regressors} on {Appa}-{Real} {Database}},
  isbn      = {978-1-5090-4023-0},
  url       = {http://ieeexplore.ieee.org/document/7961727/},
  doi       = {10.1109/FG.2017.20},
  urldate   = {2025-10-26},
  booktitle = {2017 12th {IEEE} {International} {Conference} on {Automatic}
               {Face} \& {Gesture} {Recognition} ({FG} 2017)},
  publisher = {IEEE},
  author    = {Agustsson, Eirikur and Timofte, Radu and Escalera, Sergio and
               Baro, Xavier and Guyon, Isabelle and Rothe, Rasmus},
  month     = may,
  year      = {2017},
  pages     = {87--94}
}

@inproceedings{puc_analysis_2021,
  address   = {Amsterdam, Netherlands},
  title     = {Analysis of {Race} and {Gender} {Bias} in {Deep} {Age}
               {Estimation} {Models}},
  isbn      = {978-90-827970-5-3},
  url       = {https://ieeexplore.ieee.org/document/9287219/},
  doi       = {10.23919/Eusipco47968.2020.9287219},
  urldate   = {2025-10-26},
  booktitle = {2020 28th {European} {Signal} {Processing} {Conference}
               ({EUSIPCO})},
  publisher = {IEEE},
  author    = {Puc, Andraz and Struc, Vitomir and Grm, Klemen},
  month     = jan,
  year      = {2021},
  pages     = {830--834}
}

@misc{hwang_2010_is,
  author  = {Hwang, Stephen W. and Atia, Mina and Nisenbaum, Rosane and Pare, Dwayne E. and Joordens, Steve},
  month   = {10},
  pages   = {136-141},
  title   = {Is Looking Older than One’s Actual Age a Sign of Poor Health?},
  doi     = {10.1007/s11606-010-1537-0},
  urldate = {2025-10-27},
  volume  = {26},
  year    = {2010},
  journal = {Journal of General Internal Medicine}
}

@misc{swanson_2011_objective,
  author  = {Swanson, Eric},
  month   = {09},
  pages   = {1124-1131},
  title   = {Objective assessment of change in apparent age after facial rejuvenation surgery},
  doi     = {10.1016/j.bjps.2011.04.004},
  volume  = {64},
  year    = {2011},
  journal = {Journal of Plastic, Reconstructive & Aesthetic Surgery}
}

@article{gupta_2003_customers,
  author   = {Sunil Gupta and Donald R. Lehmann},
  title    = {Customers as assets},
  journal  = {Journal of Interactive Marketing},
  volume   = {17},
  number   = {1},
  pages    = {9-24},
  year     = {2003},
  doi      = {10.1002/dir.10045},
  url      = {
              https://doi.org/10.1002/dir.10045
              },
  eprint   = {
              https://doi.org/10.1002/dir.10045
              }
}

@ARTICLE{Bontempi2025-xm,
  title     = "{FaceAge}, a deep learning system to estimate biological age
               from face photographs to improve prognostication: a model
               development and validation study",
  author    = "Bontempi, Dennis and Zalay, Osbert and Bitterman, Danielle S and
               Birkbak, Nicolai and Shyr, Derek and Haugg, Fridolin and Qian,
               Jack M and Roberts, Hannah and Perni, Subha and Prudente, Vasco
               and Pai, Suraj and Dekker, Andre and Haibe-Kains, Benjamin and
               Guthier, Christian and Balboni, Tracy and Warren, Laura and
               Krishan, Monica and Kann, Benjamin H and Swanton, Charles and De
               Ruysscher, Dirk and Mak, Raymond H and Aerts, Hugo J W L",
  journal   = "The Lancet Digital Health",
  publisher = "Elsevier",
  volume    =  7,
  number    =  6,
  month     =  jun,
  year      =  2025
}

@InProceedings{chen2014cross,
    author="Chen, Bor-Chun
    and Chen, Chu-Song
    and Hsu, Winston H.",
    editor="Fleet, David
    and Pajdla, Tomas
    and Schiele, Bernt
    and Tuytelaars, Tinne",
    title="Cross-Age Reference Coding for Age-Invariant Face Recognition and Retrieval",
    booktitle="Computer Vision -- ECCV 2014",
    year="2014",
    publisher="Springer International Publishing",
    address="Cham",
    pages="768--783",
    isbn="978-3-319-10599-4"
}

@ARTICLE{jacobs1991adaptive,
  author={Jacobs, Robert A. and Jordan, Michael I. and Nowlan, Steven J. and Hinton, Geoffrey E.},
  journal={Neural Computation}, 
  title={Adaptive Mixtures of Local Experts}, 
  year={1991},
  volume={3},
  number={1},
  pages={79-87},
  doi={10.1162/neco.1991.3.1.79}}

\end{document}